# SPOC learner's final grade prediction based on a novel sampling batch normalization embedded deep neural network method

Zhuonan Liang[1], Ziheng Liu[1], Huaze Shi[1], Yunlong Chen[1], Yanbing Cai[2], Hong Hong[3], Yating Liang[2], Yafan Feng[1], Yuqing Yang[4,5], Jing Zhang[6], Peng Fu[2] *

[1] School of Science, Nanjing University of Science and Technology, Nanjing Jiangsu 210094, China

[2] School of Computer Science and Engineering, Nanjing University of Science and Technology, Nanjing Jiangsu 210094, China

[3] Department of Electrical and Computer Engineering, University of California, Davis, Davis, CA, 95616, USA

[4] Office of International Cooperation and Exchanges, Nanjing University of Finance & Economics, Nanjing, 210046, China

[5] College of Economics and Management, Nanjing University of Aeronautics and Astronautics, Nanjing.211106, China.

[6] Jiangsu Guidgine Educational Evaluation Inc, Nanjing, 210046, China

*Corresponding Author: Peng Fu. Email: fupeng@njust.edu.cn

**Abstract:** Recent years have witnessed the rapid growth of Small Private Online Courses (SPOC) which is able to highly customized and personalized to adapt variable educational requests, in which machine learning techniques are explored to summarize and predict the learners' performance, mostly focus on the final grade. However, the problem is that the final grade of learners on SPOC is generally seriously imbalance which handicaps the training of prediction model. To solve this problem, a sampling batch normalization embedded deep neural network (SBNEDNN) method is developed in this paper. First, a combined indicator is defined to measure the distribution of the data, then a rule is established to guide the sampling process. Second, the batch normalization (BN) modified layers are embedded into full connected neural network to solve the data imbalanced problem. Experimental results with other three deep learning methods demonstrate the superiority of the proposed method.

## 1 Introduction

With the continuous development of the online education and the knowledge sharing projects, online sharing courses comes into public views. Small Private Online Courses (SPOC) is one of the most popular solutions of implementing online sharing courses. SPOC benefits the educational source sharing without the constraints of time and location. As an online education method without face-to-face, tracking the learning situation of students play a key role in SPOC, guaranteeing the education effect. Predicting student final grade of courses is a straightly approaching for investigating the teaching effectiveness. The advanced research made great contributions to precisely foresee the performance of student in the final exam. I. C. Juanatas et al. [1] presented the licensure examination performance prediction based on their academic grades using Logistic Regression. That provides a binary result liked PASS or FAIL, which is insufficient for further analysis. For detailed prediction, the support vector machine (SVM) is applied. In specialized dataset, SVM shows outstanding performance compared to current methods. While it is sensitive to the distribution of training datasets, lacking the practical generality. For traditional educational data under the massive scale, Y. Yang et al. [4] proposed an improved random forest method to comprehensively predict the grade of students in the final exams. It is effective in prediction task for less students, but difficult to handle massive educational data in SPOC. The deep learning methods are applied in the educational system with large amount students [5]. A Bayesian deep learning model with variants was proposed for grade prediction under a course-specific framework [6]. While it is only applied to the offline or the traditional school education. The investigating features of online education are different from offline but providing an approaching to model the student behavior. T. Yang et al. [7] proposed a model predicting the performance of students base on online video click stream events. While there are some obstacles in some cases of the SPOC grade prediction. The special circumstance on SPOC causes the imbalanced final grades of student which concentrate on upper score between 90 and full mark. This sort of long tail distribution causes derivation compared with the standard Gaussian distribution expected in the neural network, resulting in the negative effect on predication. To address this problem, we propose a sampling batch normalization embedded deep neural network (SBNEDNN). In proposed model, we reconstructed an indictor to measure the distribution of the dataset, guiding the following sampling operation. Moreover, we embedded the batch normalization (BN) modified layers into the multilayer perceptions, furtherly handling the training difficulties caused by imbalance educational data. Our contributions can be summarized as: 1) The paper proposes a data distribution indicator $Max\_score$ (MS) to measure the imbalance situation of education data. MS comprehensively considers the skewness index and kurtosis index of data. It is the guideline in the data processing procedure. 2) An improved imbalance data processing method is proposed in this paper. This method does not simply transform data into the uniform distribution. It reconstructs the resampling method to fit the hypothetical distribution in the following neural network design under the guideline of MS. 3) Experiments results with the widely used deep learning methods prove the superiority of our proposed method.

**2 Sampling-BN embedded deep neural network method**

In this part, we will describe the proposed model SBNEDNN. It includes two main components: data processing and neural predicting. To encounter the influence caused by distribution, a modified data process is designed, including diagnosing and shifting. For the diagnosing, a reconstructed indicator of data is applied to evaluate the distribution, integrating skewness and kurtosis rules. Then the adjustment of data using sampling is operated guided by the indicator, acting as the shifting module. After the processed, the data will be trained by the modified batch-normalized fully connected networks.

**Data Processing**
In the SBNEDNN method, we firstly evaluate the distribution of the dataset. The histogram presents qualitative analysis of the dataset distribution. However, it is not convincing to provide the quantitative analysis of the distribution, such that a define evaluation index is needed for guidance of the following sampling operation. Skewness, kurtosis, and standard deviation each measures the characteristics of data distribution. We constructed a quantitative indicator integrates three characteristics indicators above, diagnosing the data distribution and guiding the sampling. Specifically, the corresponding formulas are generated from the standard Gaussian distribution.

The skewness $S$ can indicate the extent that a given distribution varies from a normal distribution, which can be written as Equation (1):

$$S = \frac{1}{n} \sum \left[ \left( \frac{X_i - \mu}{\sigma} \right)^3 \right], \tag{1}$$

where $S$ denotes the skewness, $X_i$ represents the $i_{th}$ student's score, $\mu$ represents the mean value of the final grade of all students and $\sigma$ represents the standard variance of the scores of all students, $n$ represents the number of students.

The skewness indicates the numerical imbalanced characteristics of data distribution. In practice, when $S < 0$, the probability distribution graph is biased to the left, on the contrast, when the $S > 0$, it is biased to the right. $S = 0$ is a good symbol that the data is relatively evenly distributed on both sides of the average value. Another measure in our indicator is kurtosis $K$, which indicates the steepness of the probability distribution of a random variable. The measure can be written as Equation (2):

$$K = \frac{1}{n} \sum \left[ \left( \frac{X_i - \mu}{\sigma} \right)^4 \right] - 3, \tag{2}$$

where $K$ denotes the kurtosis, $\mu$ represents the mean value of the scores of all students, $\sigma$ represents the standard variance of the scores of all students, $n$ represents the numbers of students. Specially, it assumes that the $K$ value of the data being standard Gaussian distribution equal 0, which is the result of -3. When the kurtosis $K > 0$, it means that the data distribution is sharper than the normal distribution, on the other hand, $K < 0$ means that it is squat compared to the normal distribution. By combining $S$ and $K$, $\sigma$, the developed indicator $Max\_score$ can be written as Equation (3):

$$MS = \frac{\max(abs(S, K))}{\sigma}, \tag{3}$$

where $max(\cdot)$ means selecting the largest data value, $abs(\cdot)$ represents calculating the absolute value, $S$ represents the skewness, $K$ represents the kurtosis, $\sigma$ represents the standard variance of $MS$. The proposed indicator presents how the data converge to Gaussian distribution considering both skewness and kurtosis. Moreover, we introduced the hypothesis testing to evaluate the data distribution via measuring the $MS$. Z-test formula is opted as the criteria in this part. The formula can be presented as:

$$Z = \frac{MS}{\sigma}, \tag{4}$$

where $\sigma$ presents the standard value of $MS$. $Z$ is the testing statistical value in the hypothesis test. $Z$ reflects the difference between testing data distribution and Gaussian distribution, considering skewness and kurtosis through selecting the most distant indicator. To exemplify, the skewness and kurtosis of one specified dataset are 0.194 and 0.373, and $\sigma$ in $Z$ is obviously a value less than the standard values of skewness and kurtosis, representing as $M$ which is 0.36 in this case. $Z$ is expected ranging from 0 to 1.96 as statistical standard Z-test threshold value $\varepsilon$, while it is 1.036 in this example. According to the hypothesis testing result, it can be supposed that the data distribution obeys the standard Gaussian distribution.

When we obtain the $S$ and $K$, $Max\_score$ of the dataset of scores, we take $S$ and $K$ to zero as the direction of sampling and stop sampling until the value of $Z$ is less than $\varepsilon$ (usually set as 1.96). For better illustration, a flow chart is exhibited in Figure 1. Under the guidance of the $Max\_score$, the classes with more samples are first processed by an under-sampling approach, the classes with less samples are first processed by using an over-sampling approach, which helps to balance class distribution by replicating data of minority sample. Figure 2 shows these two methods detailly.

Once the data is processed by above processes, the fully connected neural network with BN modification is applied in the subsequent prediction, the three steps of the proposed SBNEDNN model can be summarized as Table 1. For a better illustration, the pre-processing procedure is represented by Figure 3.a, while the deep learning model prediction is displayed on the Figure 3.b and Figure 3.c.

**Table 1**: The implementation process of the SBNEDNN Method

---

**SBNEDNN Method**

**Step1:** Data evaluation and preprocess.
do
{

$$S = \frac{1}{n} \sum \left[ \left( \frac{X_i - \mu}{\sigma} \right)^3 \right];$$

$$K = \frac{1}{n} \sum \left[ \left( \frac{X_i - \mu}{\sigma} \right)^4 \right] - 3;$$

$$Max\_score = \frac{\max(abs(S,\ K))}{\sigma};$$

$$Z = \frac{Max\_socre}{\sigma};$$

Procedure *sampling*
{
    if ($S < 0$){
        *oversampling* lower classes;
    }else{
        *oversampling* upper classes;
    }
    if ($K > 0$){
        *undersampling* medium classes;
    }else{
        *oversampling* medium classes;
    }
}
} while ($Z > \varepsilon$)

**Step2**: Score prediction
Split the training dataset into mini batches;
*Loop* {
    *Loop* (3){
        *fully connected layer* (mini batch);
        *batch normalization layer* (y);
    }
}

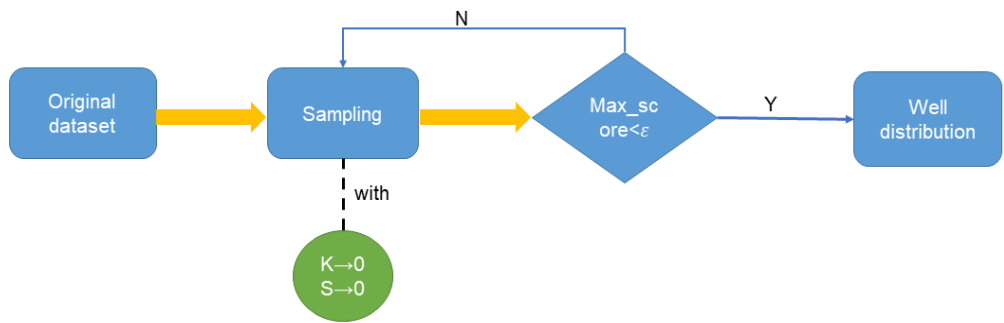

**Figure 1**: The logistic of data preprocess. $K$ presents the kurtosis and $S$ presents the skewness.

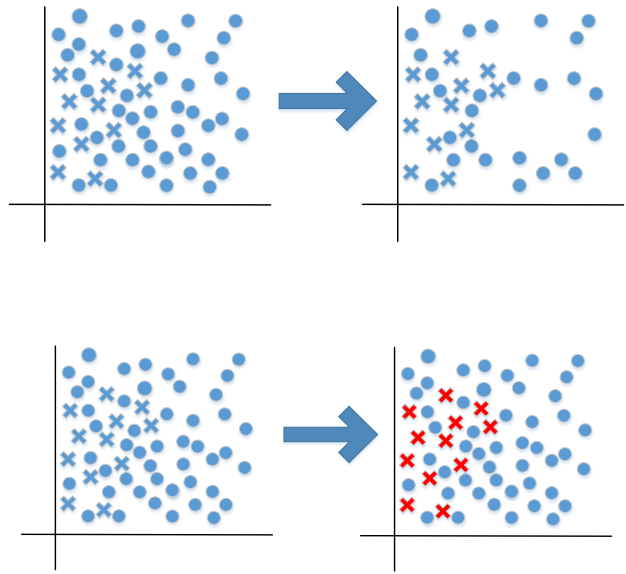

**Figure 2**: The changes of data distribution from original to well-balanced after under sampling and over sampling

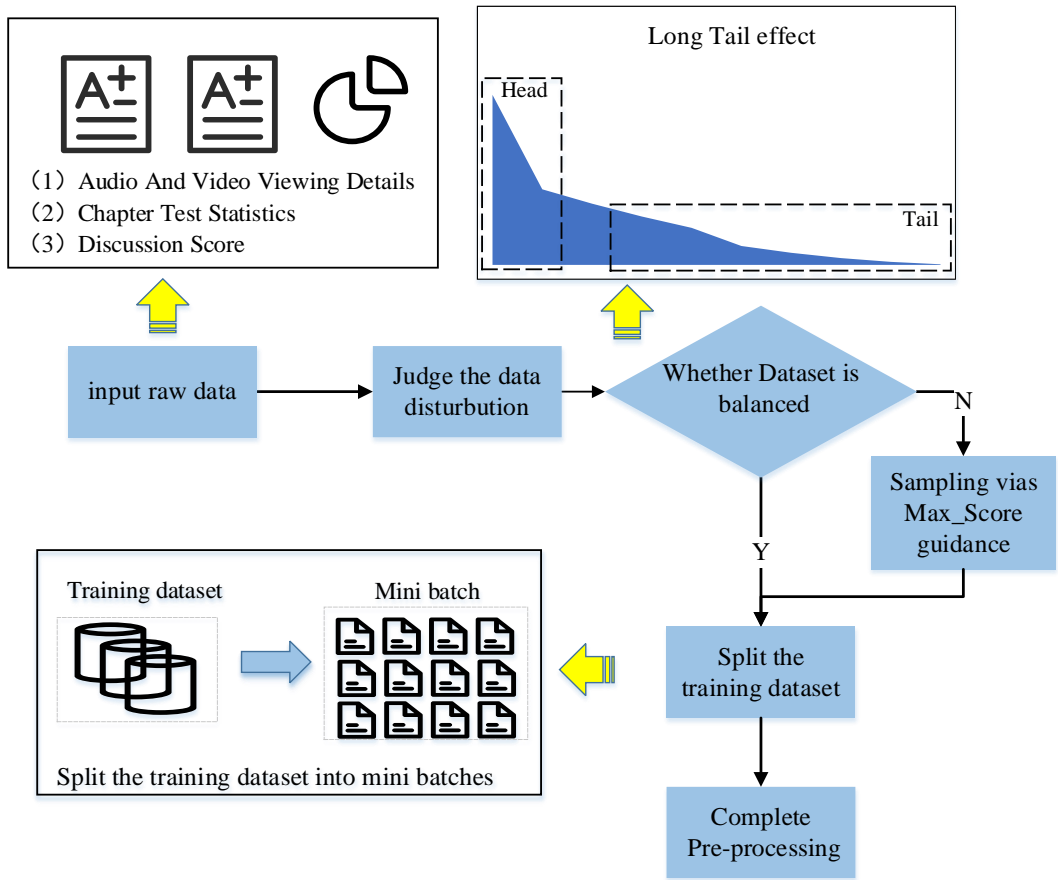

**Figure 3.a:** The Pre-processing procedure of SBNEDNN Method

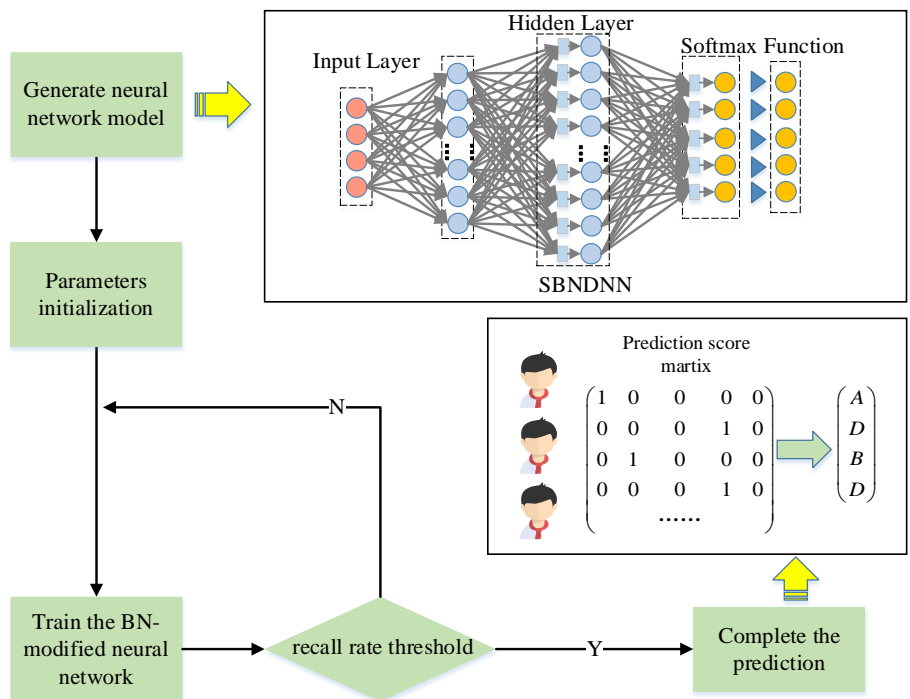

**Figure 3.b:** The model training procedure of SBNEDNN Method

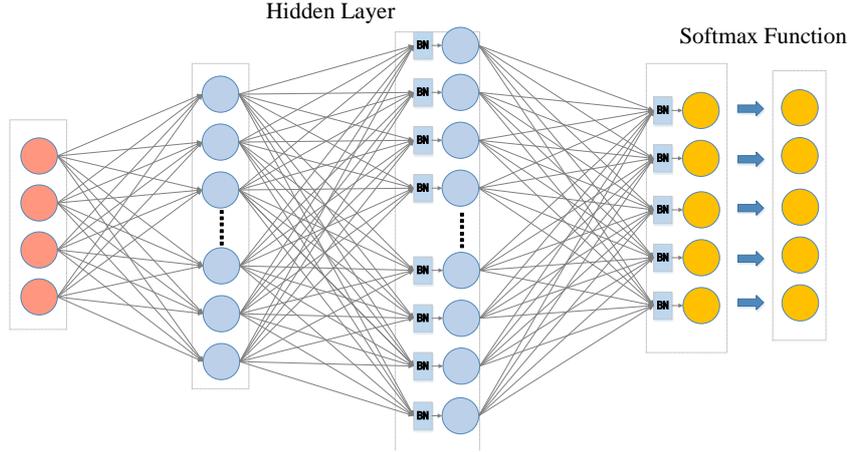

**Figure 3.c**: The training network of SBNEDNN

## 3 Experiments

### *3.1 Dataset*

We conduct the experiments on the dataset Medic 2020 and Statistics 2020. Medic 2020 is from a real-world SPOC for medical graduate students on Chaoxing platform[1]. Statistics 2020 is the desensitization learning record for the year 2020 on the "Medical Statistics Program of the National Association of Medical College Graduate Schools SPOC Platform". It tracks the learner's performance of exercises and tests during their studies on the online learning platform. The utilized SPOC dataset contains various features about students, including student background, learning behavior record, and test performance. To focus the learning-related features, we simplify the dataset, filtering out the background-related features, like age, hometown, and gender. Therefore, we focus on the features corresponding to the learners to predict their final grades. The filtered dataset has sixty-nine features, which can be classified into three categories, "Audio and Video", "Chapter Test Record", "Discussion Score". The detailed description of these features is presented in Table 2. In the SPOC dataset, some records of the features are missing in some specified students. They were dropped in the data preprocess. After processing above, the final dataset totally contains 40,494 sets of student learning record, and each set has sixty-nine features. To facilitate the prediction, we classify the grades into six groups according to Equation (4) for alleviate the imbalance issues in dataset:

$$level = \begin{cases} L1 & 0 \leq grade \leq 70 \\ L2 & 70 \leq grade \leq 80 \\ L3 & 80 \leq grade \leq 90 \\ L4 & 90 \leq grade \leq 93 \\ L5 & 93 \leq grade \leq 95 \\ L6 & 95 \leq grade \leq 100 \end{cases}, \quad (4)$$

where L1 to L6 denote six classes. It is noting that the features remain the original presentation, a number range from 0 to 100. From Figure 5 we can see that the class L6 has the most students, while the class L4 owns the least students. This is because in SPOC learning, students are prone to get great final grades if they work hard enough. Specifically, the train-test split ratio is 70 versing 30.

**Table 2**: Features Description

| Three main feature classes | Description |
|---|---|
| Audio and Video (60 features) | The score of behavior in sixty lesson sections. |
| Chapter Test Statistics (8 features) | The score of each chapter test. |
| Discussion Score (1 features) | The score in the discussion activities. |

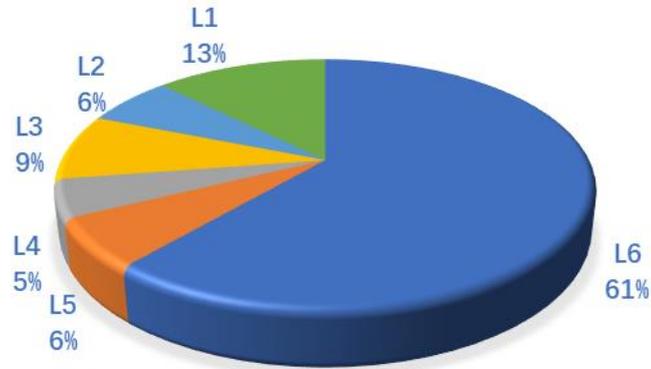

**Figure 5**: The number of samples in each class with the experimental dataset.

### 3.2 Parameter setting and evaluation metrics

To initiate the model, we opt the Xavier initialization and generate the parameter from the gauss distribution. The average value and variance are calculated from the training dataset. The mini-batch size is 128. Adam algorithm is applied for optimization. The activation function is Sigmoid. To measure the performance of different structures, we selected the prediction accuracy to evaluate whether model can predict the final grade of students.

We implement the all experiments with PyTorch via Python and conduct them on a Linux server with two 2.0GHz Intel Xeon E5-2683 CPUs and two GeForce 1080Ti GPUs.

### 3.3 Experimental results

#### 3.3.1 Baseline experiments comparison

To show the superiority of the SBNEDNN method, three widely used deep learning methods, including CNN, RNN, LSTM, are applied with the pre-balanced dataset. The prediction accuracy of these methods is shown in Table 3. The SBNEDNN method achieves the highest prediction accuracy compared to the other three methods in overall. Moreover, we can get further interesting information if taking educational practice into consideration. As Table 3 shown, the proposed model shows the superior performance among the medium three classes. According to the educational experience, the students of the medium classes ranging 70 to 90 should be specially focused for promoting them or preventing them from dropping. Recognizing them and taking the proper tutorial are significant in the SPOC education. From this aspect, LSTM and our method are suitable in the situation above, while LSTM are less accurate.

#### 3.3.2 Model structure comparison

In this section, the prediction performance of different model structures will be compared and discussed. It is divided into position of BN layers and count of fully connected layer. Through these result and relative discussion, it demonstrates that the structure of our model is comprehensive and properly constructed for accurate prediction.

To find the suitable BN layer position, we implemented three positional variants of model in the experiments. The result is shown as Table 4. It is no doubt that SBNEDNN achieves the excellent prediction according to the experimental result, while the network structure without BN layers gets the bottom of experiment as expected. Different positions of BN layer influence the transformation of data to Gaussian distribution, avoiding gradient disappearing and distribution deviation during the propagation. The result also convinces that the BN operation earlier means more positive effect in the network.

Moreover, the amount of fully connected layer plays significant role in the deep learning prediction model. We implemented the modal variants with different multiple fully connected layers. Specifically, all the variant models are BN-embedded. In experiments, we focus on the increasing rate of training time and accuracy, which presents how much time cost in the unit accuracy promotion. Experimental result is presented as Table 5.a and Table 5.b. It demonstrates that more fully connected layers enhance the network accuracy, while it also causes the longer training time and lower adoptability. The result also shows that considerable prediction does not require such deep network which cost much more time. Moreover, the over deep network structure leads to overfitting and data deviation, causing negative effect in the educational predication task. It requires the prediction model with generality and adoptability due to handling the various student behavior features in practical situation.

**Table 3.a**: Experiments of SBNEDNN and three deep learning methods on Medic 2020

| Methods | Acc of L1 | Acc of L2 | Acc of L3 | Acc of L4 | Acc of L5 | Acc of L6 | Tol acc |
|---|---|---|---|---|---|---|---|
| LSTM | 93.06% | 100% | 97.33% | 91.92% | 95.65% | 82.47% | 92.74% |
| RNN | 94.53% | 96.15% | 88.47% | 88.68% | 95.21% | 78.59% | 91.49% |
| CNN | 97.01% | 97.11% | 98.84% | 98.49% | 97.82% | 84.87% | 95.55% |
| **SBNEDNN** | 95.83% | 100% | 100% | 100% | 95.65% | 80.41% | **98.85%** |

**Table 3.b**: Experiments of SBNEDNN and three deep learning methods on Statistics 2020

| Methods | Acc of L1 | Acc of L2 | Acc of L3 | Acc of L4 | Acc of L5 | Acc of L6 | Tol acc |
|---|---|---|---|---|---|---|---|
| LSTM | 84.50% | 83.66% | 96.97% | 89.94% | 92.77% | 52.33% | 87.75% |
| RNN | 79.84% | 94.77% | 94.55% | 98.32% | 87.35% | 69.19% | 86.78% |
| CNN | 87.60% | 92.81% | 96.36% | 96.09% | 92.77% | 79.07% | 83.63% |
| **SBNEDNN** | 96.90% | 94.12% | 100% | 94.41% | 98.80% | 80.23% | **92.84%** |

Table 4: The accuracies with different BN layouts

| Structure | Layers | Medic 2020 | Statistics 2020 |
|---|---|---|---|
| Structure 1 | Dense -> Dense -> Dense | 67.68% | 51.86% |
| Structure 2 | Dense -> Dense -> BN -> Dense | 94.83% | 62.75% |
| Structure 3 | Dense -> BN -> Dense -> Dense | 97.47% | 64.21% |
| **SBNEDNN** | **Dense -> BN -> Dense -> BN -> Dense** | **98.85%** | **92.84%** |

Table 5: Accuracy and training time of three different structures

| Layer depth | Medic 2020 | Statistics 2020 |
|---|---|---|
| 3 | 98.85% | 92.84% |
| 4 | 98.92% | 93.87% |
| 5 | 98.99% | 95.85% |
| 6 | 96.74% | 94.60% |
| 7 | 94.52% | 91.18% |

## 4 Conclusion

This paper constructs SBNEDNN method for the prediction of SPOC learner's grade. First, an indicator is defined to measure the distribution of dataset as well as guide the sampling process. Second, a BN modified neural network is built to train the data after sampling. Then, a SBNEDNN method is developed to improve the prediction accuracy with imbalance data. Experiment results with the comparison of other three widely used deep learning methods show the effectiveness and supremacy of our method.

**Declarations**
The authors have no relevant financial or non-financial interests to disclose.

**Acknowledgments**
We would like to thank Jiangsu Guidgine Educational Evaluation Inc. and Alliance of Graduate School, Medical University in China for providing SPOC data.

**Funding Statement:**
This work was in part supported by the Undergraduate Student Out-of-class Academic Foundation of Nanjing University of Science and Technology, in part supported by the National Natural Science Foundation of China under Grant no. 61801222, and in part supported by the Fundamental Research Funds for the Central Universities under Grant no. 30919011230.

**Conflicts of Interest:**
The authors declare that they have no conflicts of interest to report regarding the present study.